\documentclass[10pt, journal, cspaper]{IEEEtran}

\usepackage{cite}

\usepackage{subimages}
\usepackage{subfig}
\setfigdir{figs}

\usepackage[cmex10]{amsmath}
\usepackage{amsthm}
\usepackage{amssymb}

\newcommand{\T}{^\mathsf{T}}

\usepackage[ruled,lined,linesnumbered]{algorithm2e}

\usepackage[breaklinks=true,colorlinks=true]{hyperref}

\begin{document}
%
\title{Feature Learning by Multidimensional Scaling and its Applications in Object Recognition}

\newif\iffinal
\finaltrue
\newcommand{\jemsid}{113568}


\iffinal
  \author{%
    \IEEEauthorblockN{Quan Wang \qquad Kim L. Boyer\\}
    \IEEEauthorblockA{%
      Signal Analysis and Machine Perception Laboratory\\
      Department of Electrical, Computer, and Systems Engineering\\
      Rensselaer Polytechnic Institute, Troy, NY 12180, USA\\
      \href{mailto:wangq10@rpi.edu}{wangq10@rpi.edu} \qquad
      			\href{mailto:kim@ecse.rpi.edu}{kim@ecse.rpi.edu}
      }
  }
\else
  \author{Sibgrapi paper ID: \jemsid \\ }
\fi

%


\maketitle

\begin{abstract}

We present the MDS feature learning framework, 
in which multidimensional scaling (MDS) is applied on high-level pairwise image distances to learn 
fixed-length vector representations of images. 
The aspects of the images that are captured by the learned features, which we call MDS features, 
completely depend on what kind of image distance measurement is employed. 
With properly selected semantics-sensitive image distances, the MDS features provide rich semantic information about the images that is not captured by 
other feature extraction techniques. 
In our work, we introduce the iterated Levenberg-Marquardt algorithm for solving MDS, and study the MDS feature learning with IMage Euclidean Distance (IMED) and Spatial Pyramid Matching (SPM) distance. We present experiments on both synthetic data and real images --- the 
publicly accessible UIUC car image dataset. The MDS features based on SPM distance achieve exceptional performance for the car recognition task. 
%
\end{abstract}


\IEEEpeerreviewmaketitle


\section{Introduction}
To represent an image by a fixed-length feature vector, there are generally two groups of approaches, often referred to as hand-designed features and feature learning, respectively. 
In this section, we briefly review several commonly used methods from each group, and relate the proposed MDS feature learning to these existing methods. 

\subsection{Hand-Designed Features}
Most hand-designed features, or sometimes called hand-crafted features, focus on capturing the color, texture and gradient information in the image. Generally, these features have a closed form to be computed, without looking at other images. Some popular yet simple hand-designed image features include color-histogram, wavelet transform coefficients \cite{wavelet}, scale-invariant feature transform (SIFT) \cite{sift}, color-SIFT \cite{color-sift}, speeded up robust features (SURF) \cite{surf}, histogram of oriented gradients (HOG) \cite{HOG} and local binary patterns (LBP) \cite{lbp}. 
To represent an image with one fixed-length feature vector, there are generally three ways: 
(1) First, these features can be computed for the entire image, but the resulting feature vector will fail to embed the spatial relationship between different objects or different locations in the image. (2) Second, the image can be first uniformly divided into $M \times N$ blocks. Then these features can be computed for each block, and can be concatenated to make a long feature vector. (3) Further, the division of the image does not have to be uniform, but can be arbitrary. We can just put random rectangular or circular masks onto the image, and compute features for each mask (or ``patch''), then concatenate. To do this, the division must be consistent for all images. 

The divide-and-concatenate methods will result in very large feature vectors. Given a large dataset, PCA can be used to reduce the dimensionality. 

\subsection{Feature Learning}
Feature learning has often been used as a synonym of deep learning, especially in recent years, and often refers to 
recent techniques such as sparse coding \cite{nature, sparse-coding}, auto-encoder \cite{denoising_autoencoder}, convolutional neural networks \cite{ConvolutionalNN}, 
restricted Boltzmann machines \cite{science}, and deep Boltzmann machines \cite{dbm}. 
However, we believe this interpretation of feature learning is literally imprecise. 
Feature learning should be more generally defined as the opposite to hand-designed features --- 
it should refer to any technique that learns a fixed-length vector representation of each image in the dataset by 
utilizing the pattern distribution of the entire dataset, or optimizing a target function that is defined on the entire dataset. Any technique that can generate a feature representation of each image without looking at the entire dataset should fail to fall into this category. 

We further categorize existing feature learning methods into two subgroups: feature learning with raw intensities, and feature learning with hand-designed features. The proposed MDS feature learning falls into a third new subgroup: feature learning with image distance measurement.  

\subsubsection{Feature Learning with Raw Intensities}

This subgroup of methods treat the feature learning problem as a dimensionality reduction problem, 
where the original high-dimensional data are the image intensities, either gray-level or RGB values. 
Efforts on data dimensionality reduction have a long history \cite{dim_review}, 
dating from the early work on PCA \cite{PCA} and its nonlinear form, kernel PCA \cite{kPCA}, 
to the recent work on sparse coding and deep learning \cite{nature, science, sparse-coding, ConvolutionalNN, denoising_autoencoder, dbm}. 
In all these methods, high dimensional data, such as an image, is represented by a low dimensional vector.  
Each entry of this vector describes one salient varying pattern of all images within the training set.  

Assume we have a dataset $\mathbf{X}=(\mathbf{x}_1, \mathbf{x}_2, \dots , \mathbf{x}_N)\T$, where each $\mathbf{x}_i$ ($1 \leq i \leq N$) is one data point. We briefly review several dimensionality 
reduction methods below. 
\begin{itemize}
\item
\textbf{PCA} linearly projects vector $\mathbf{x}$ to $\mathbf{y}=\mathbf{Ax}$, where $\mathbf{A}$ is obtained by performing eigenvector decomposition on the covariance matrix $\mathbf{S_x=XX}\T$. 
\item
\textbf{Kernel PCA} first constructs a kernel matrix $\mathbf{K}$, where 
each entry of this matrix is obtained by evaluating the 
kernel function $k(\cdot , \cdot)$ on two data points: 
\begin{equation}
\mathbf{K}_{i,j}=k(\mathbf{x}_i,\mathbf{x}_j) . 
\end{equation}
Then the Gram matrix is constructed as 
\begin{equation}
\mathbf{\widetilde K} = \mathbf{K} - \mathbf{1}_N \mathbf{K} - \mathbf{K} \mathbf{1}_N + \mathbf{1}_N \mathbf{K} \mathbf{1}_N , 
\end{equation}
where $\mathbf{1}_N$ is the $N \times N$ matrix with all elements equal to $1/N$. 
Next the eigenvector decomposition problem $\mathbf{\widetilde{K}a}_l=\lambda_l N \mathbf{a}_l$ is solved ($\mathbf{a}_l$ is eigenvector and $\lambda_l$ is eigenvalue) and the projected vector $\mathbf{y}$ is computed by 
\begin{equation}
y_l=\sum \limits _{i=1}^N a_{li} k(\mathbf{x,x}_i). 
\end{equation}
\item
\textbf{Auto-encoders}  first normalize all $\mathbf{x}_i$'s to $[0,1]$, and map them to 
$\mathbf{y}_i=s(\mathbf{Wx}_i+\mathbf{b})$, where $s(\cdot)$ is a sigmoid function. 
A reconstruction is computed by $\mathbf{z}_i=s(\mathbf{W'y}_i+\mathbf{b'})$. The weight matrices $\mathbf{W}$ and $\mathbf{W'}$, and the bias vectors $\mathbf{b}$ and $\mathbf{b'}$ are obtained by minimizing the average reconstruction error, which can be defined as either traditional square error or cross-entropy. 
\end{itemize}

In PCA and kernel PCA, different entries of $\mathbf{y}$ correspond to eigenvectors of different importance, while in auto-encoder, they are equivalently important. 
 
These techniques have been shown effective on problems such as face recognition \cite{eigenface, eigenface2} and even concept recognition \cite{google-cat}. 
However, most of these methods 
require all input data to have exactly the same size. If the input is an image, then the image has to be cropped and resized to be consistent with other images in the dataset. However, cropping the image means loss of information, and resizing the image means change of aspect ratio, which will result in distorted object shapes. 

\subsubsection{Feature Learning with Hand-Designed Features}
\label{sec:bov}
One popular method that falls into this subgroup is the bag-of-visual-words (BOV) method \cite{bov1,bov2,feifei}. 
This method first divides the image into local patches or segments the image into distinct regions, and then extracts hand-designed features for each patch/region. Rather than being concatenated, these feature vectors make an unordered set, or also referred to as ``bag''. 
By performing clustering on the union of all those unordered sets for all training images, 
a visual vocabulary is established. 
Now the set of feature vectors previously extracted from each image can be transformed into a ``word-frequency'' histogram by simply counting which cluster (visual word) is assigned to each patch/region. 
The ``word-frequency'' histogram can be optionally normalized to generate the final fixed-length vector. 

One extension of BOV is the Fisher Vector (FV) method \cite{fv1,fv2}. Rather than simply counting the word frequency, which can be viewed as the 0-order statistics, FV encodes higher order statistics (up to the second order) about the distribution of 
local descriptors assigned to each visual word. Another extension is the Spatial Pyramid Matching \cite{spm}, which gives different weights to features in different image division levels, and defines an image similarity measurement using the pyramid matching kernel \cite{pyramid_kernel}. 

\section{Method}
In this section, we first review the basics of MDS and its existing solutions, and then introduce our 
own solution --- the iterated Levenberg-Marquardt algorithm (ILMA). Next, we discuss and compare some popular image distance measurement techniques in recent literature.  

\subsection{Multidimensional Scaling: Problem Definition}
As a statistical technique for the analysis of data similarity
or dissimilarity, multidimensional scaling (MDS) has been well applied to areas such as information visualization  \cite{MDS_book} and surface flattening \cite{flattening1, flattening2}. 
Here we briefly review the basic concepts and definitions of MDS. For convenience, we will use the word ``image'' instead of ``data'' or ``object'' in the context, but we keep in mind that MDS is a technique for general purposes. 

Suppose we have a set of $N$ images $\Omega=\lbrace I_1, I_2, ..., I_N \rbrace$, and there is a distance measurement $d(I_i,I_j)$ defined between each pair of images $I_i$ and $I_j$. 
Note that $d: \Omega \times \Omega \rightarrow \mathbb{R}_{\geq 0}$ is only a measurement of image dissimilarity, not 
necessarily a metric on set $\Omega$ in the strict sense, since the subadditivity triangle inequality does not 
necessarily hold. 
Multidimensional scaling is the problem of representing each image $I_i \in \Omega$ by a point (vector) in a low dimensional space $\mathbf{x}_i \in \mathbb{R}^m$, such that the interpoint Euclidean distance in some sense approximates the distance between the corresponding images \cite{MDS}. In Section \ref{sec:image_distance} we will discuss how to define the image distance/dissimilarity measurement. Here we focus on the mathematical definitions related to MDS. 

For a pair of images $I_i$ and $I_j$, let their low dimensional ($m$-d) representations be $\mathbf{x}_i$ and $\mathbf{x}_j$. The representation error is defined as $e_{ij}=d(I_i,I_j)-||\mathbf{x}_i-\mathbf{x}_j||$, where $||\cdot||$ denotes the $L^2$-norm. The \textit{raw stress} is defined as the sum-of-squares of the representation errors: 
\begin{equation}
\textrm{Stress}^*=\sum\limits_{1 \leq i < j \leq N}e_{ij}^2 , 
\end{equation}
while the \textit{normalized stress} (also known as \textit{Stress-1}) is defined as 
\begin{equation}
\textrm{Stress-1}=\sqrt{\dfrac{\sum\limits_{1 \leq i < j \leq N}e_{ij}^2}{\sum\limits_{1 \leq i < j \leq N}||\mathbf{x}_i-\mathbf{x}_j||^2}} .
\end{equation} 
MDS models require the interpoint Euclidean distances to be ``as equal as possible'' to the image distances. Thus we can either minimize the raw stress or normalized stress.  
We compactly represent the image distances by an $N \times N$ symmetric matrix 
$\mathbf{D}=\left[ d(I_i,I_j) \right] _{N \times N}$ 
with all diagonal values equal to $0$, 
and represent the low dimensional vectors by an $N \times m$ matrix 
$\mathbf{X}=(\mathbf{x}_1, \dots, \mathbf{x}_N)\T$. 
Using the raw stress as the loss function, the MDS problem can be stated as:
\begin{equation}
\label{eq:min_raw_stress}
\mathbf{X}^*=\arg  \min\limits_{\mathbf{X}} \sum\limits_{1 \leq i < j \leq N}
\left( d(I_i,I_j)-||\mathbf{x}_i-\mathbf{x}_j|| \right) ^2 . 
\end{equation}

\subsection{Solutions for Multidimensional Scaling}
There are lots of existing methods for solving Eq. (\ref{eq:min_raw_stress}), such as 
Kruskal's iterative steepest descent approach \cite{MDS} 
and de Leeuw's iterative majorization algorithm (SMACOF) \cite{SMACOF}. 
In 2002, Williams demonstrated the connection between kernel PCA and metric MDS \cite{connection}, thus metric MDS problems can also be solved by solving kernel PCA. 

In our work, we introduce an iterative least squares solution to the MDS optimization problem. 
We note that 
in Eq. (\ref{eq:min_raw_stress}), the raw stress is minimized with respect to $\mathbf{X}$, which has $N \times m$ entries in total. Thus, when $N$ is large, this nonlinear optimization problem becomes computationally intractable if we attempt to solve for all entries in one step. 
Inspired by the iterated conditional modes (ICM) method \cite{ICM}, which was developed to solve Markov random fields (MRF), we introduce the two-stage \emph{iterated Levenberg-Marquardt algorithm (ILMA)}. The basic idea of this algorithm is 
to repeatedly minimize the raw stress with respect to one $\mathbf{x}_i$ while holding all other $\mathbf{x}_i$'s fixed. 
For this purpose, we maintain a constraining set of the indices of the $\mathbf{x}_i$'s to be fixed. 
In the initialization stage, indices of all images are selected into the constraining set in a random order. 
In the adjustment stage, we repeatedly adjust all $\mathbf{x}_i$'s in a randomly permuted order. 
By doing so, each time we only need to minimize the raw stress with respect to $m$ variables, instead of 
$N \times m$, which greatly reduces the complexity of the problem. 
The subproblem can be viewed as a least squares problem, and can be solved by the standard 
Levenberg-Marquardt algorithm \cite{levenberg, marquardt}. 
Since the total raw stress is monotonically non-increasing through time, the convergence of the 
adjustment is guaranteed. 
The details of the two-stage algorithm are given in Algorithm \ref{alg:Levenberg-Marquardt}. We will call the low dimensional vectors $\{ \mathbf{x}_i \}$ as \emph{MDS features} or \emph{MDS codes} in the context. 

One advantage of our method is that we provide a unified framework for both MDS model training and new data encoding.  
In MDS model training, pairwise image distances are measured within the training set $\Omega_{\mathrm{train}}$, and Algorithm \ref{alg:Levenberg-Marquardt} is applied to encode each training image $I_i \in \Omega_{\mathrm{train}}$ to its MDS code $\mathbf{x}_i$. 
Now given a new image $\widetilde{I}$, we measure the distance from this image to all training images  $d(\widetilde{I},I_i)$, and find its MDS code $\widetilde{\mathbf{x}}$ by:  
\begin{eqnarray}
\label{eq:encoding}
\min\limits_{\widetilde{\mathbf{x}}} \sum\limits_{I_i \in \Omega_{\mathrm{train}}} ( ||\widetilde{\mathbf{x}}-\mathbf{x}_{i}|| - 
d(\widetilde{I},I_i) ) ^2 , 
\end{eqnarray} 
which can be directly solved as a least squares problem using the standard 
Levenberg-Marquardt algorithm. 
We follow this practice for the training and testing of MDS models in the experiment in Section \ref{sec:car}.  

\begin{algorithm}[h!]
 \SetKwInOut{Input}{input}\SetKwInOut{Output}{output}
 \Input{Distance matrix $\mathbf{D}=[ D_{i,j} ]$, $1 \leq i,j \leq N$;\\
  Max number of iterations $M$;}
 \Output{MDS codes $\mathbf{X}=(\mathbf{x}_1, \dots, \mathbf{x}_N)\T$;}
 \BlankLine
 \textbf{begin} initialization stage:\\
 Randomly choose a non-diagonal entry $D_{i_0,j_0}$ in $\mathbf{D}$;\\
 Set $\mathbf{x}_{i_0}=(0, 0, \dots, 0)\T$;\\
 Set $\mathbf{x}_{j_0}=(D_{i_0,j_0}, 0, \dots, 0)\T$;\\
 Initialize the constraining set $A=\lbrace i_0, j_0 \rbrace$;\\
 \While{$|A|<N$}
 {
	Randomly choose $j^* \notin A$;\\ 
	Use the standard Levenberg-Marquardt algorithm to find the $\mathbf{x}_{j^*}$ that minimizes:
	$$\sum\limits_{i \in A} ( ||\mathbf{x}_{j^*}-\mathbf{x}_{i}|| - D_{j^*,i} ) ^2 ;$$\\
	Add $j^*$ to $A$, set $\mathbf{x}_{j^*}$ in $\mathbf{X}$;
 }
 \textbf{begin} adjustment stage:\\
 \For{$t\leftarrow 1$ \KwTo $M$}
 {
 	Generate a random permutation $(p_1, p_2, \dots, p_N)$ of integers $1$ to $N$;\\
 	\For{$s\leftarrow 1$ \KwTo $N$}
 	{
		Take $p_{s}$ out of $A$;\\ 
		Use the standard Levenberg-Marquardt algorithm to find the $\mathbf{x}_{p_s}$ that minimizes:
		$$\sum\limits_{i \in A} ( ||\mathbf{x}_{p_s}-\mathbf{x}_{i}|| - D_{p_s,i} ) ^2 ;$$\\
	Add $p_s$ back to $A$, update $\mathbf{x}_{p_s}$ in $\mathbf{X}$;
	}
	\lIf{$\Delta \mathrm{Stress}^* / \mathrm{Stress}^*< \epsilon$}
	{break;}
 }
 \textbf{done}\\
 \caption{The two-stage iterated Levenberg-Marquardt algorithm (ILMA).}
 \label{alg:Levenberg-Marquardt}
\end{algorithm}

\subsection{Image Distance Measurement}
\label{sec:image_distance}
The measurement of the similarity or dissimilarity between two images is of essential significance in content-based image retrieval \cite{cbir_survey1, cbir_survey2}. There are some very simple forms of image distances, such as the traditional Euclidean distance on raw image intensities, 
and the earth mover's distance (EMD) on image color histograms \cite{EMD}. 
Here, we briefly describe two popular image distance measurement methods: the IMage Euclidean Distance (IMED) \cite{IMED} and 
the Spatial Pyramid Matching (SPM) distance \cite{spm}. These distances will be evaluated in our experiment on real images in Section \ref{sec:car}. 

\subsubsection{IMED}
The IMED is a generalized form of the traditional Euclidean distance on raw image intensities. Give two gray-level images $I_1$ and $I_2$ of the same size, the traditional Euclidean distance is defined as the square root of the sum-of-squares of intensity difference at each corresponding image location:
\begin{equation}
\label{eq:euclidean_distance}
d_{\mathrm{Euclidean}}^2(I_1,I_2)=\sum\limits_{(r,c)} \left( I_1^{(r,c)}-I_2^{(r,c)} \right)^2, 
\end{equation}
where $I_1^{(r,c)}$ denotes the intensity at row $r$ and column $c$ in image $I_1$. 
In contrast, IMED also counts for the intensity difference at different locations, but assigns a weight to it, which is a function of the Euclidean distance of the two locations: 
\begin{eqnarray}
\label{eq:IMED}
d_{\mathrm{IMED}}^2(I_1,I_2)=\sum\limits_{(r,c)}\sum\limits_{(r',c')} 
\left( I_1^{(r,c)}-I_2^{(r,c)} \right) \cdot \nonumber \\ 
g(r,c,r',c') \cdot \left( I_1^{(r',c')}-I_2^{(r',c')} \right),
\end{eqnarray}
where 
\begin{equation}
\label{eq:IMED_g}
g(r,c,r',c')=f(\sqrt{(r-r')^2+(c-c')^2}), 
\end{equation}
and $f(\cdot)$ is a continuous monotonically decreasing function, usually the Gaussian function. 
An interesting observation by Wang \emph{et al.} \cite{IMED} is that the IMED (\ref{eq:IMED}) on two images is equivalent to the traditional Euclidean distance (\ref{eq:euclidean_distance}) on a blurred version of the two images. The blur operation is called standardizing transform (ST) by the authors. 

Although IMED has shown promising performance on some recognition experiments in \cite{IMED}, we can see that it is still a low-level image distance measurement, based on the raw intensities, without embedding any semantic information. Another disadvantage of IMED is that it is only defined on images of the same size. 
We will apply MDS on IMED distances for the experiment in Section \ref{sec:car}, where we 
use Gaussian function for $f(\cdot)$ in Eq. (\ref{eq:IMED_g}) and set $\sigma=1$, 
and we call this method IMED-MDS. 

\subsubsection{SPM Distance}
\label{sec:spm}
The spatial pyramid matching (SPM) \cite{spm} is based on Grauman and Darrell's work on pyramid matching kernel \cite{pyramid_kernel}, which measures the similarity of two sets of feature vectors by partitioning the feature space on different levels and taking the sum of weighted histogram intersection functions. Lazebnik \textit{et al.}'s spatial pyramid matching is an ``orthogonal'' approach --- it performs pyramid matching in the 2-d image space, and uses $k$-means for clustering in the feature space (edge points and SIFT features). With a visual vocabulary of size $M$ (number of clusters), and $L$ partition levels, spatial pyramid vectors of dimensionality $M\frac{1}{3}(4^{L+1}-1)$ are generated, and spatial pyramid matching similarities $K^L(I_i,I_j)$ between images $I_i$ and $I_j$ are measured.  
Authors of \cite{spm} recommend parameter setting of $M=200$ and $L=2$. 

The similarity value $K^L(I_i,I_j)$ lies in $[0,1]$, where 1 is for most similar, and 0 for least similar. We have many ways to define image distances using the similarities, such as: 
\begin{eqnarray}
\label{eq:spm1}
d_{\mathrm{SPM1}}(I_i,I_j)&=&1-K^L(I_i,I_j) , \\
\label{eq:spm2}
d_{\mathrm{SPM2}}(I_i,I_j)&=&- \ln ((1-\epsilon)K^L(I_i,I_j)+\epsilon)  ,
\end{eqnarray}
where $\epsilon$ is a small value. We set $\epsilon=0.001$ in (\ref{eq:spm2}) for our experiment in Section \ref{sec:car}. 

Unlike IMED, SPM distance is based on hand-designed features such as SIFT and edge points, instead of raw intensities. It models the spatial co-occurrence of different feature clusters, and thus is more semantics-sensitive. Besides, SPM distance does not require the size of images to be the same. We will apply MDS on the two SPM distances defined by Eq. (\ref{eq:spm1}) and Eq. (\ref{eq:spm2}), and we call them SPM1-MDS and SPM2-MDS, respectively.

\section{Experiments}
\label{sec:experiments}
We present two experiments. The first one is on synthetic data, and is to evaluate the running time performance of different MDS algorithms, and to compare different initialization strategies of our iterated Levenberg-Marquardt algorithm. The second one is a real image object recognition task, in which we compare MDS features with PCA features and kernel PCA features. 
In the second experiment, we use the UIUC car dataset\footnote{\url{http://cogcomp.cs.illinois.edu/Data/Car/}}, and follow a five-fold cross validation to report the classification precision and recall under different feature dimensions. 

\subsection{Synthetic Data Experiment}
\begin{figure}[h!]
  \centering
    \includegraphics[width=0.4\textwidth]{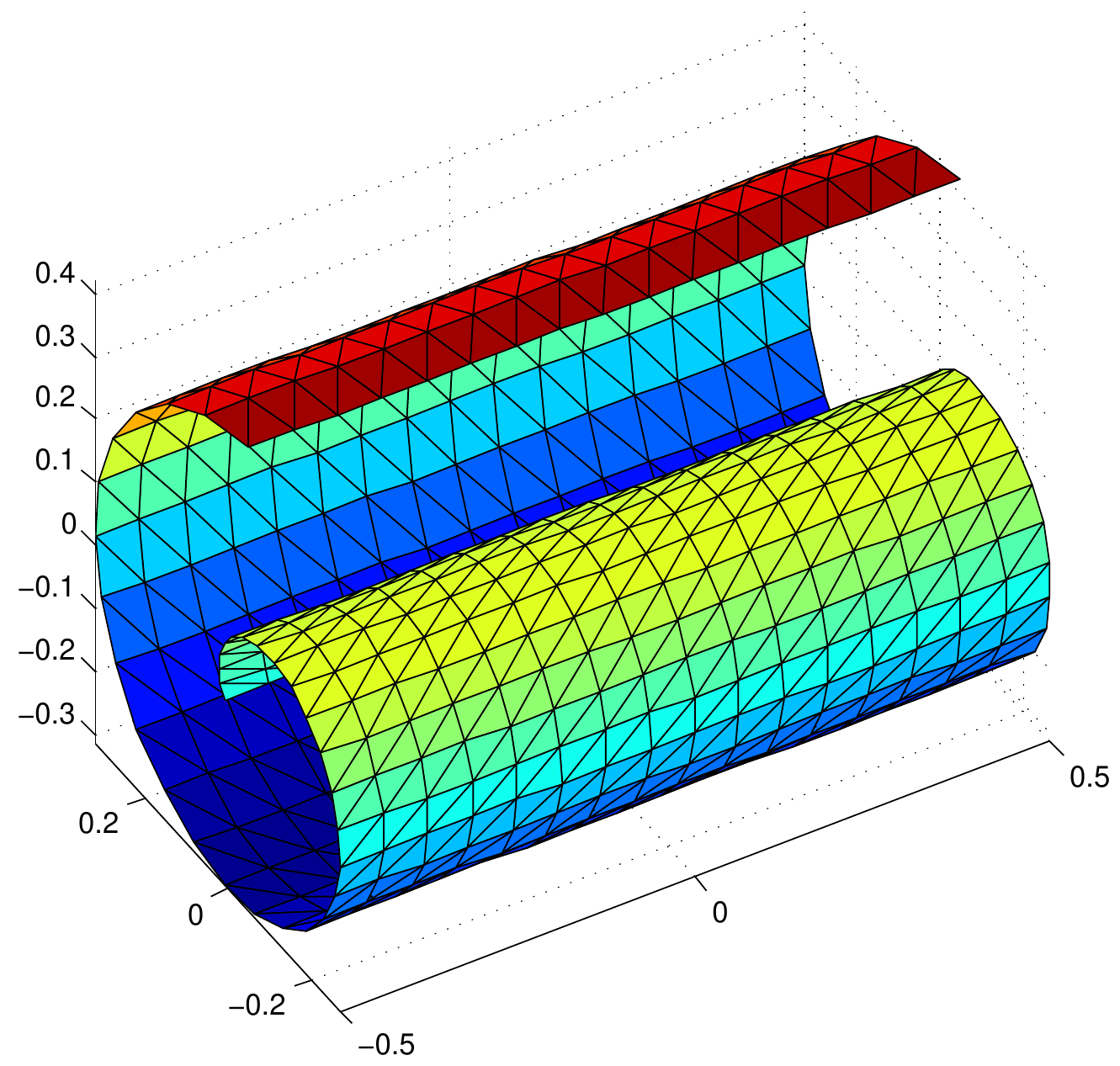}
  \caption{The Swiss roll surface with 591 points. }
  \label{fig:swiss}
\end{figure}

In this experiment, we use MDS for curved surface flattening \cite{flattening1} on the manually created \emph{Swiss roll} data, which was introduced in \cite{LLE}, and is known to be complicated due to the highly non-linear and non-Euclidean structure \cite{SMACOF_multigrid}. The Swiss roll surface contains $591$ points in $\mathbb{R}^3$, as shown in Fig. \ref{fig:swiss}. We measure the pairwise interpoint geodesic distances to construct a $591 \times 591$ distance matrix, and re-embed the Swiss roll surface into $\mathbb{R}^3$ by applying MDS on the geodesic distance matrix.  

\subsubsection{Running Time}
\begin{figure}
  \centering
    \includegraphics[width=0.47\textwidth]{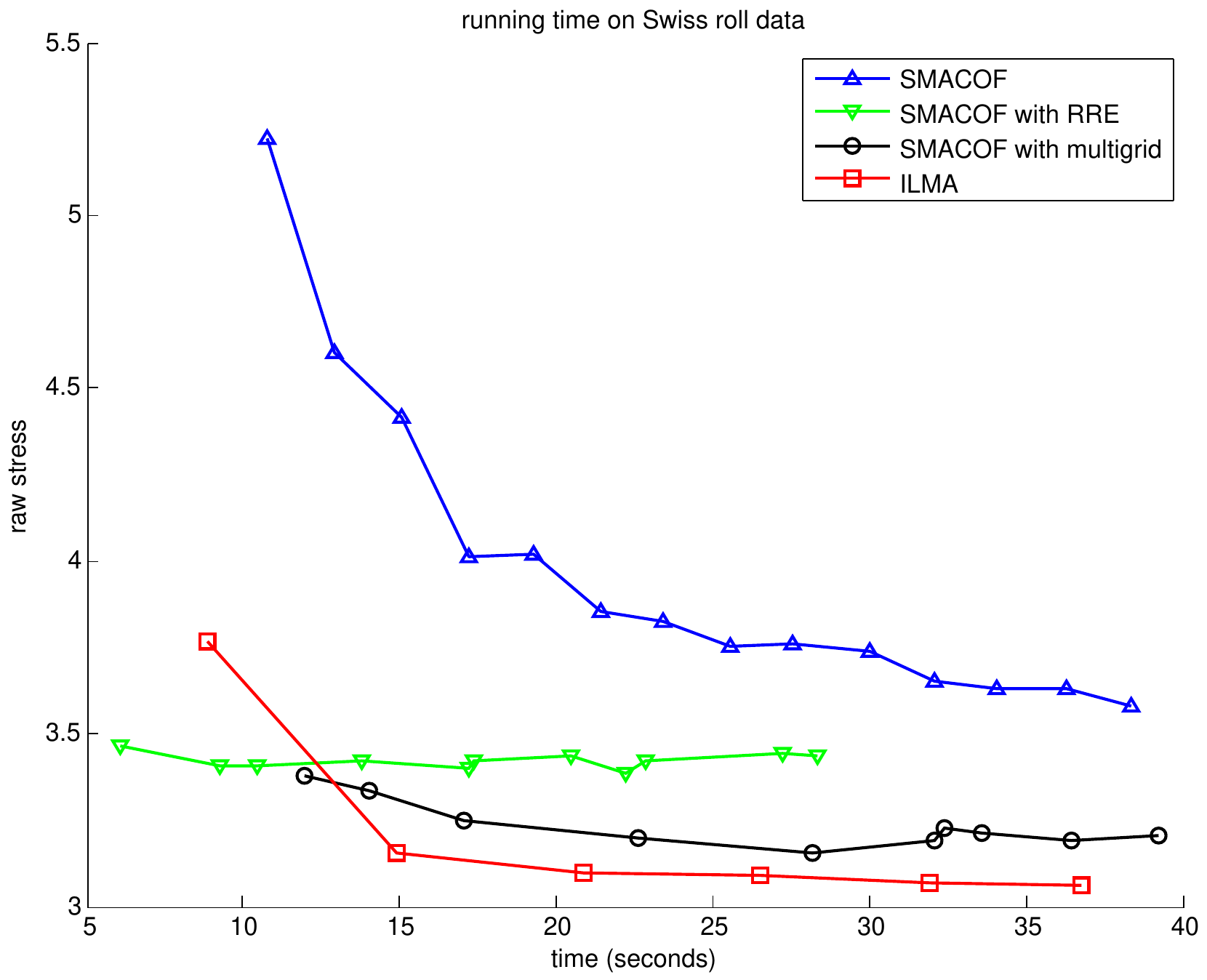}
  \caption{The raw stress \textit{vs.} running time plot of the SMACOF algorithm, its variants, and the proposed iterated Levenberg-Marquardt algorithm (ILMA) on the Swiss roll geodesic distance matrix. }
  \label{fig:ILMA_runningtime}
\end{figure}

\begin{figure*}
  \centering
    \subfloat[after initialization]
      {
       \includegraphics[height=0.33\textwidth]{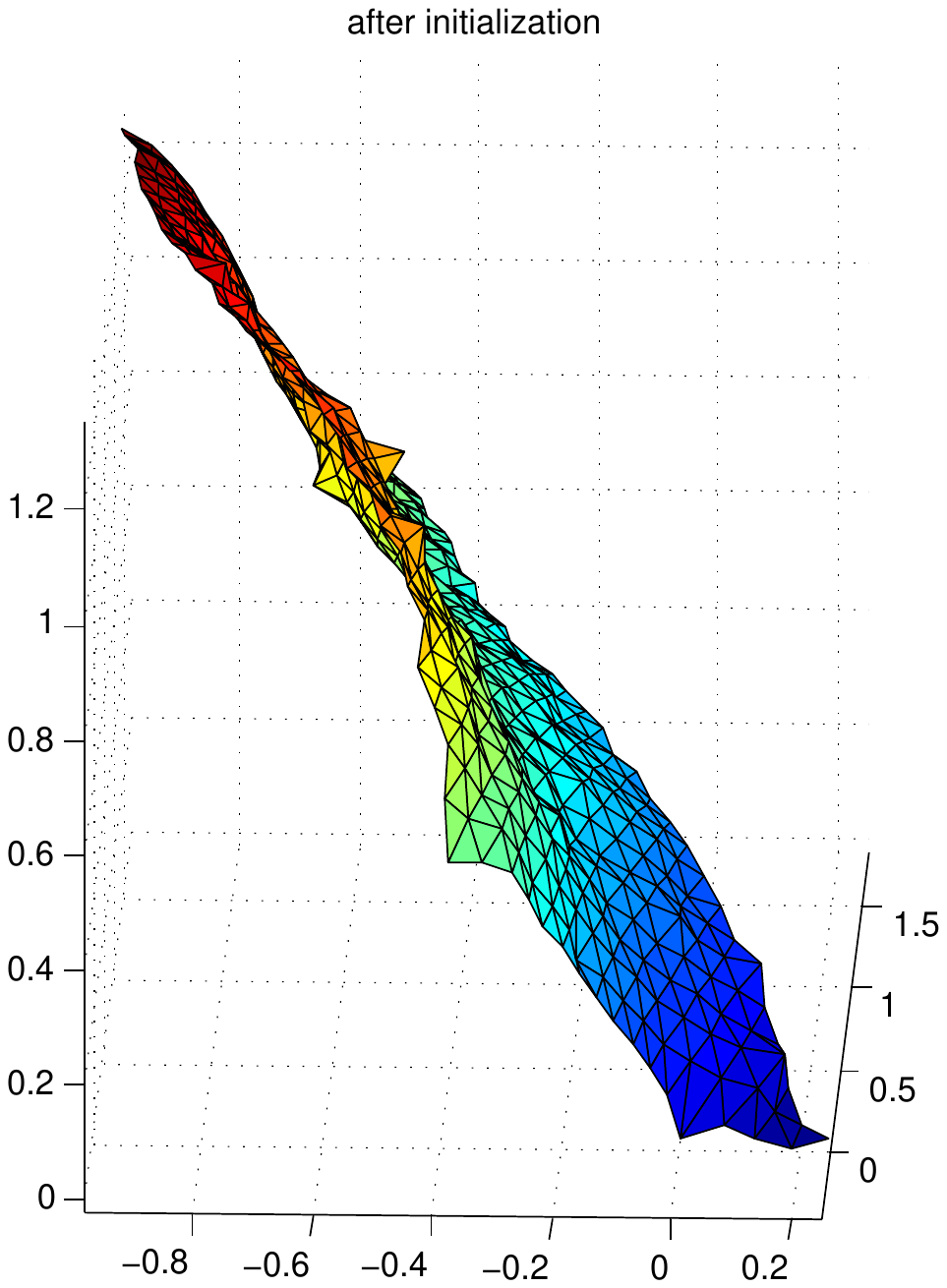}} 
    \subfloat[after 1 adjustment iteration]
      {
       \includegraphics[height=0.33\textwidth]{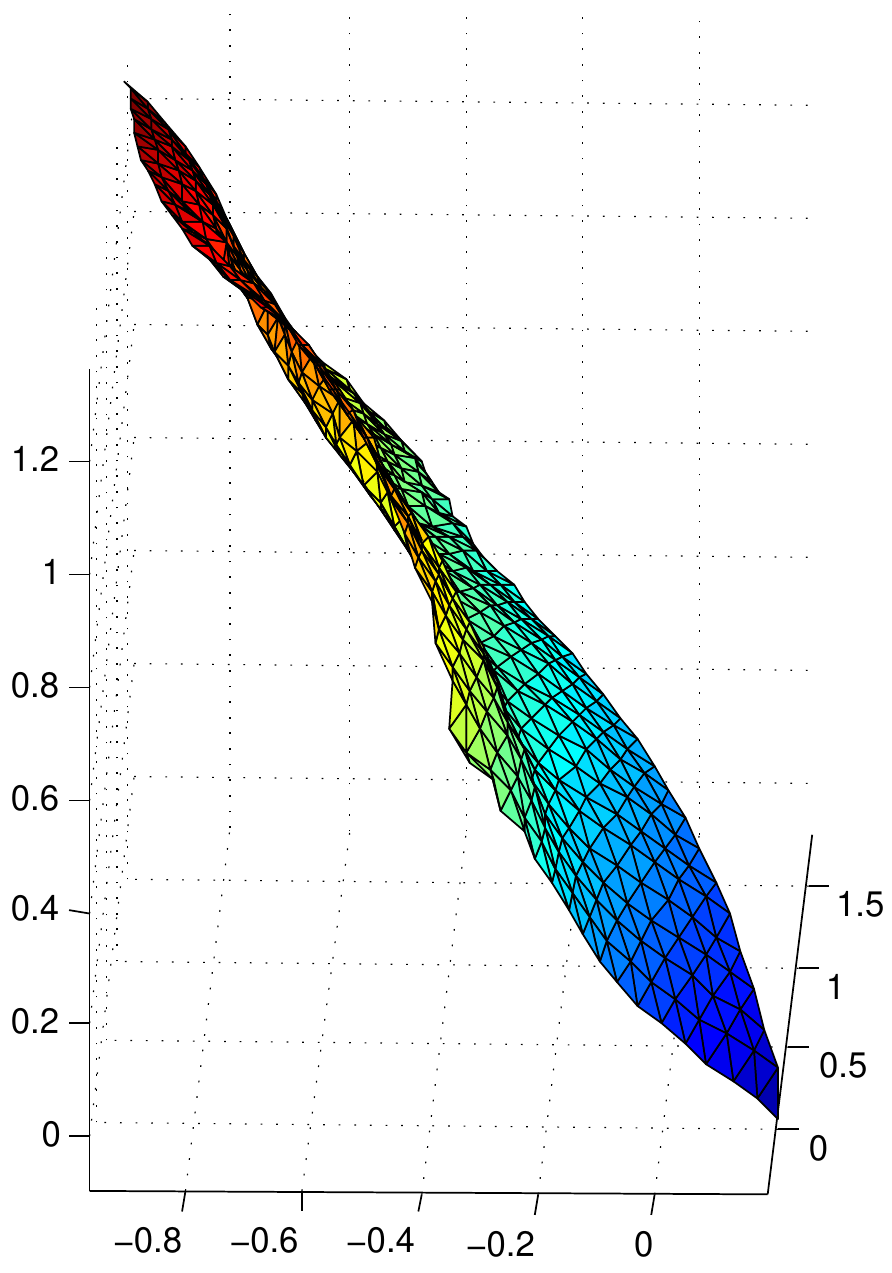}}
    \subfloat[after 2 adjustment iterations]
      {
       \includegraphics[height=0.33\textwidth]{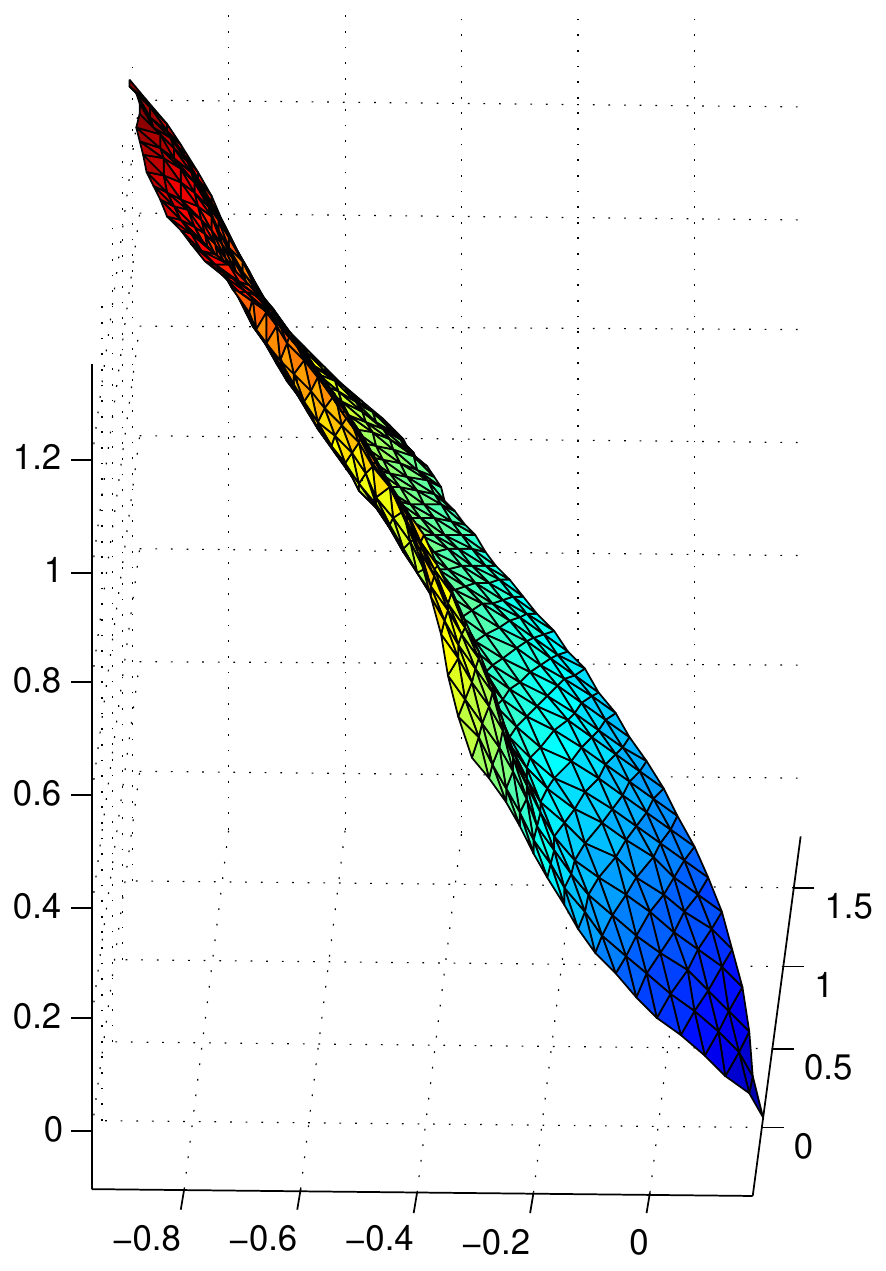}} 
    \subfloat[after 10 adjustment iterations]
      {
       \includegraphics[height=0.33\textwidth]{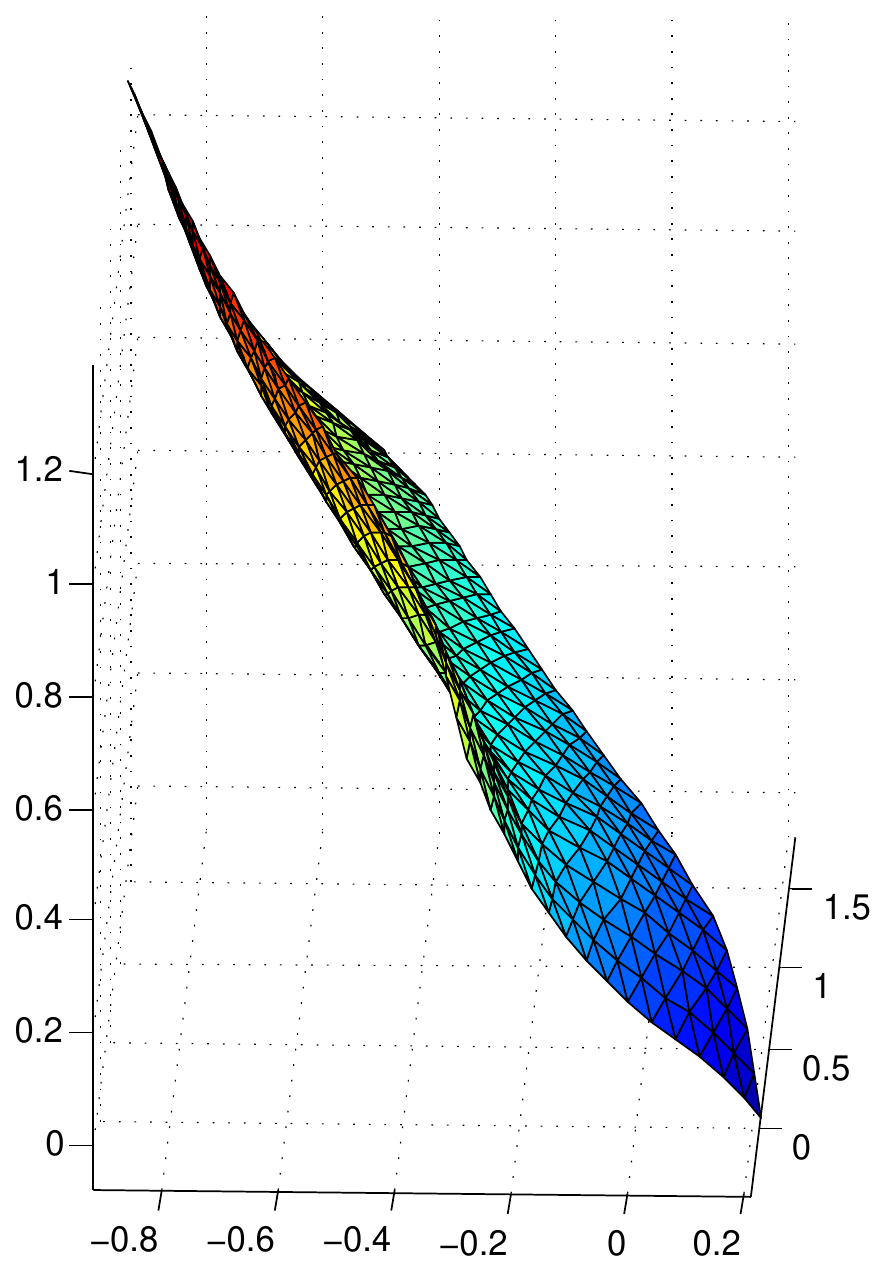}}
  \caption{Flattened Swiss roll surface by applying MDS with iterated Levenberg-Marquardt algorithm on the geodesic distance matrix. }
  \label{fig:ILMA_swiss}
\end{figure*}

First, we would like to evaluate the running time performance of the proposed iterated Levenberg-Marquardt algorithm and compare with Bronstein's implementation of the SMACOF algorithm and its variants, including SMACOF with reduced rank extrapolation (RRE) and SMACOF with multigrid \cite{SMACOF_multigrid, SMACOF_multigrid2, SMACOF_multigrid3, SMACOF_multigrid4}. 
The results are given in Fig. \ref{fig:ILMA_runningtime}, where each number in this plot is averaged on 20 independent repeated experiments, and the running time is reported on a Mac Pro with 2 $\times$ 2.4 GHz Quad-Core Intel Xeon CPU.  From Fig. \ref{fig:ILMA_runningtime}, we can see that our ILMA is an efficient solution, which runs faster and converges to a smaller raw stress value than other methods. The unrolled surfaces by ILMA in different iterations are shown in Fig. \ref{fig:ILMA_swiss}. 

\subsubsection{Initialization Strategies}
Further, we study some modifications to Algorithm \ref{alg:Levenberg-Marquardt}. The original algorithm uses a \textbf{random order strategy} in the initialization stage, but we can modify it to:
\begin{itemize}
\item \textbf{Largest-distance-first strategy:} For Algorithm \ref{alg:Levenberg-Marquardt}, in line 2 we choose the largest non-diagonal entry $D_{i_0,j_0}$ in $\mathbf{D}$ instead of a random one; in line 7, we find the $i^* \in A$ and $j^* \notin A$ that maximize $D_{i^*,j^*}$ rather than a random $j^* \notin A$. 
\item \textbf{Smallest-distance-first strategy:} For Algorithm \ref{alg:Levenberg-Marquardt}, in line 2 we choose the smallest non-diagonal entry $D_{i_0,j_0}$ in $\mathbf{D}$; in line 7, we find the $i^* \in A$ and $j^* \notin A$ that minimize $D_{i^*,j^*}$. 
\end{itemize}
If we assume that the data to be encoded are comprised of clusters, then an intuitive interpretation of the largest-distance-first strategy is that representatives of each cluster are first encoded, and they are expected to be scattered in the multidimensional space; similarly, the smallest-distance-first strategy encodes all data in one cluster first, and then moves to the nearest cluster. 

\begin{figure}
  \centering
    \includegraphics[width=0.47\textwidth]{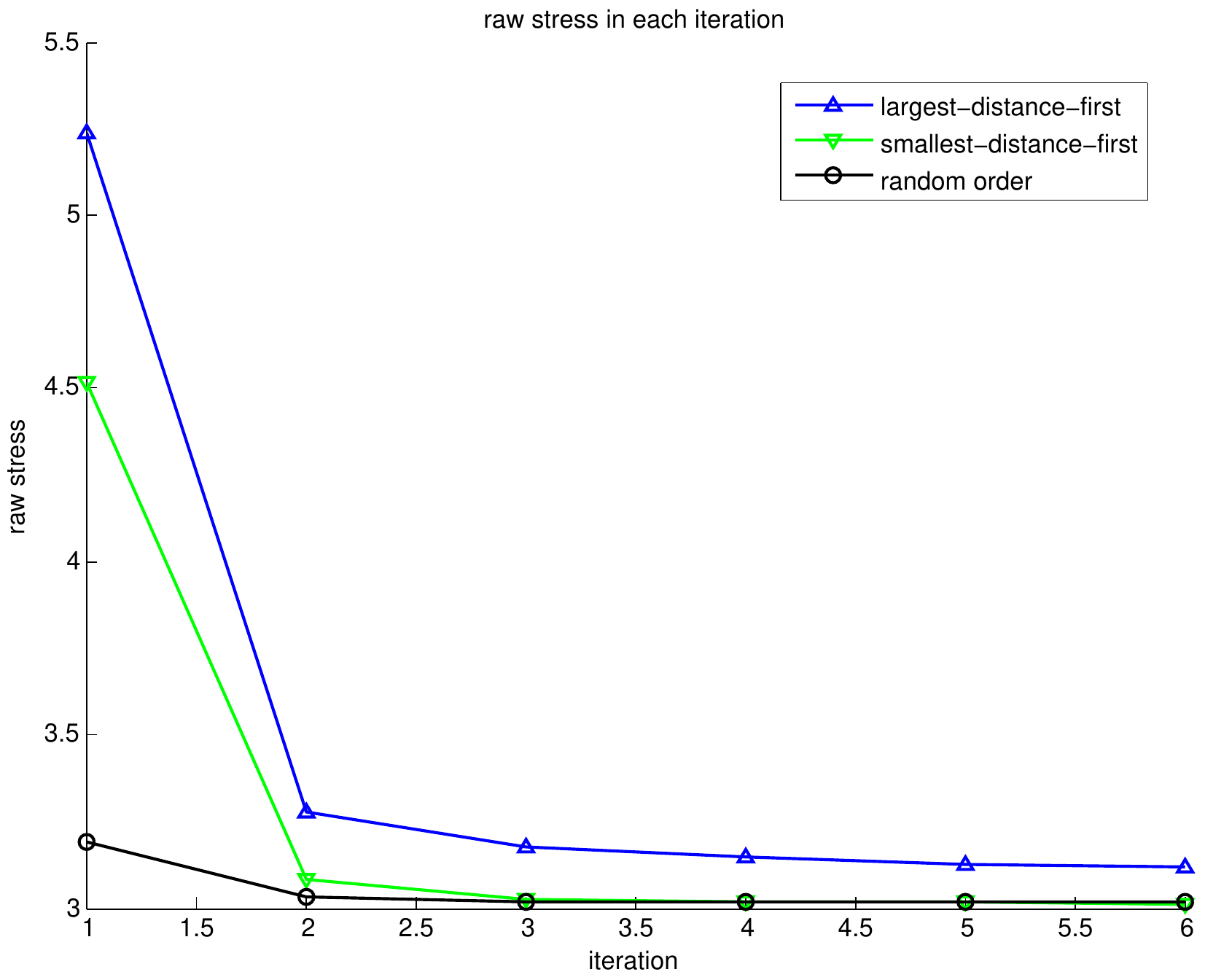}
  \caption{The raw stress in each iteration of the iterated Levenberg-Marquardt algorithm with different initialization strategies. }
  \label{fig:init_strategy}
\end{figure}

We have been using the three initialization strategies to solve the MDS problem on the Swiss roll geodesic distance matrix, and it turns out that the random order strategy converges faster than the other two, as shown in Fig. \ref{fig:init_strategy}. Again, each number in this plot is averaged on 20 independent repeated experiments. 

\subsection{Car Recognition Experiment}
\label{sec:car}
\begin{figure}
  \centering
    \includegraphics[width=0.47\textwidth]{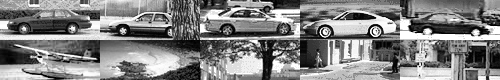}
  \caption{Example car (first row) and non-car (second row) images in UIUC car image dataset.}
  \label{fig:car_data}
\end{figure}

Now we would like to compare the performance of MDS features to the most standard and popular dimensionality reduction algorithms --- PCA \cite{PCA} and kernel PCA \cite{kPCA} on raw pixel intensities. We use the UIUC car image dataset \cite{UIUC-car}, which contains 550 car and 500 non-car gray-level images of size $40\times100$ (Fig. \ref{fig:car_data}). We can observe that all car images are side-view images, but can be either side, and can be partly occluded. We divide the total of 1050 images into five subsets, each containing 110 car images and 100 non-car images, and each time we use four subsets as training set and one as testing set. We use the following methods to generate fixed-length feature vectors for the images: 
\begin{enumerate}
  \item \textbf{PCA} We represent each $40\times100$ gray-level image by a 4000-d vector, and perform standard PCA on such vectors of the training set to get eigenvectors and low dimensional representations of the training images. Then we use the eigenvectors to get the low dimensional representations of the testing images. 
  \item \textbf{kPCA Gaussian} Similar to the above method, but we use Gaussian kernel PCA instead of standard PCA. We follow the automatic parameter selection strategy in \cite{quan_kPCA} to determine the $\sigma$. 
  \item \textbf{kPCA poly} Similar to the above two methods, but we use third-order polynomial kernel PCA instead of standard PCA. 
  \item \textbf{IMED-MDS} We first measure the IMED between each pair of training images, and run Algorithm \ref{alg:Levenberg-Marquardt} to learn the low dimensional MDS features of each training image. Then we measure the IMED from each testing image to each training image, and solve Eq. (\ref{eq:encoding}) to obtain the MDS features of each testing image.
  \item \textbf{SPM1-MDS} Similar to the above method, but we use SPM1 distance   (\ref{eq:spm1}), instead of IMED, where the SPM parameters are $M=200$ and $L=2$.   
  \item \textbf{SPM2-MDS} Similar to the above method, but we use SPM2 distance  (\ref{eq:spm2}). 
  \item \textbf{pyramid PCA} Instead of computing MDS features from SPM distances, we can also directly perform PCA on the obtained $M\frac{1}{3}(4^{L+1}-1)$-dimensional spatial pyramid vectors without measuring similarities. In our experiment, we set $M=200$ and $L=2$, and the spatial pyramid vectors are 4200-d. Evaluating this method will allow us to observe whether the MDS on SPM distance  measurement captures semantics beyond the spatial pyramids. 
\end{enumerate} 

\begin{figure*}
  \centering
    \subfloat[precision]
      {\label{fig:UIUC_precision}
       \includegraphics[width=0.32\textwidth]{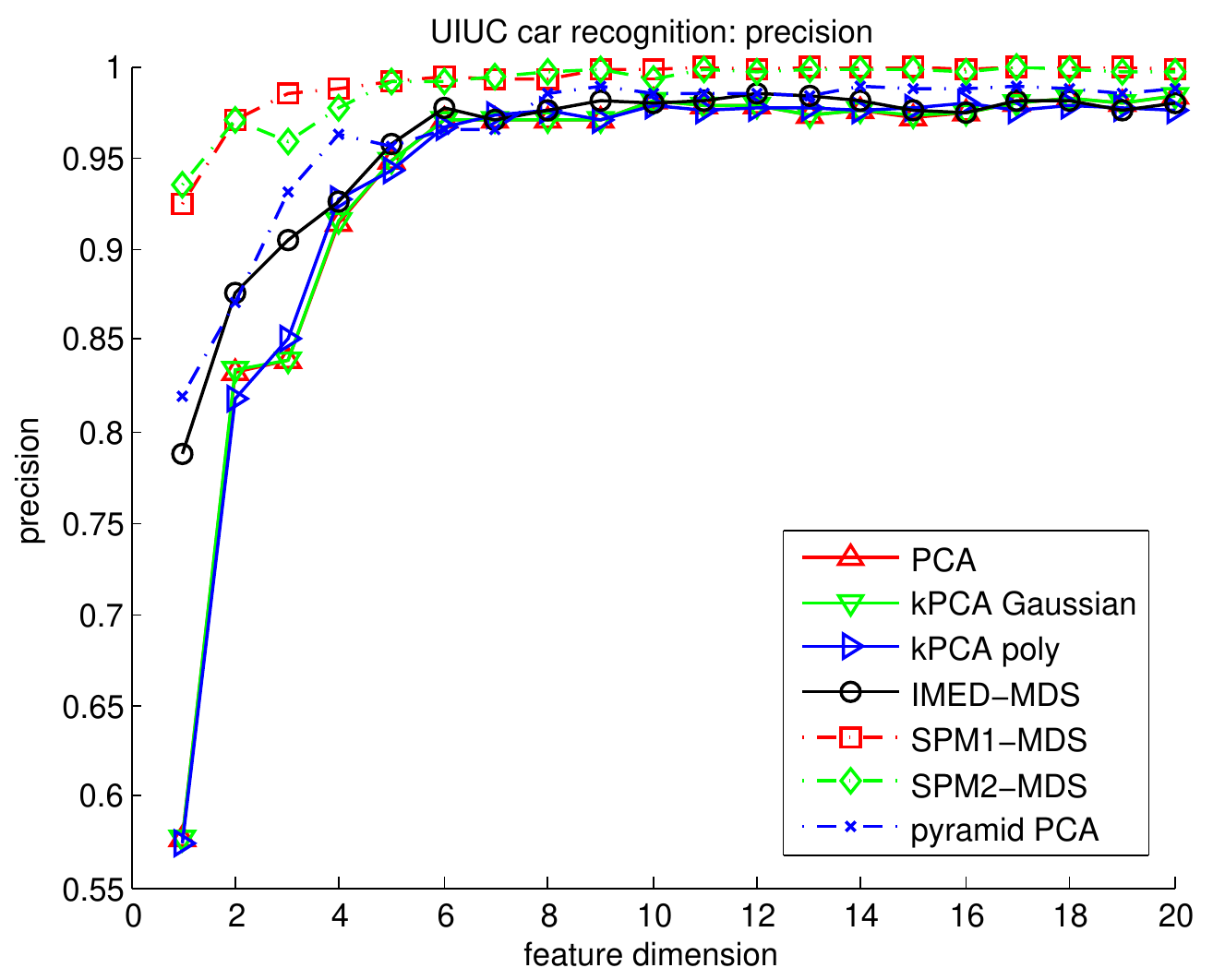}} 
    \subfloat[recall]
      {\label{fig:UIUC_recall}
       \includegraphics[width=0.32\textwidth]{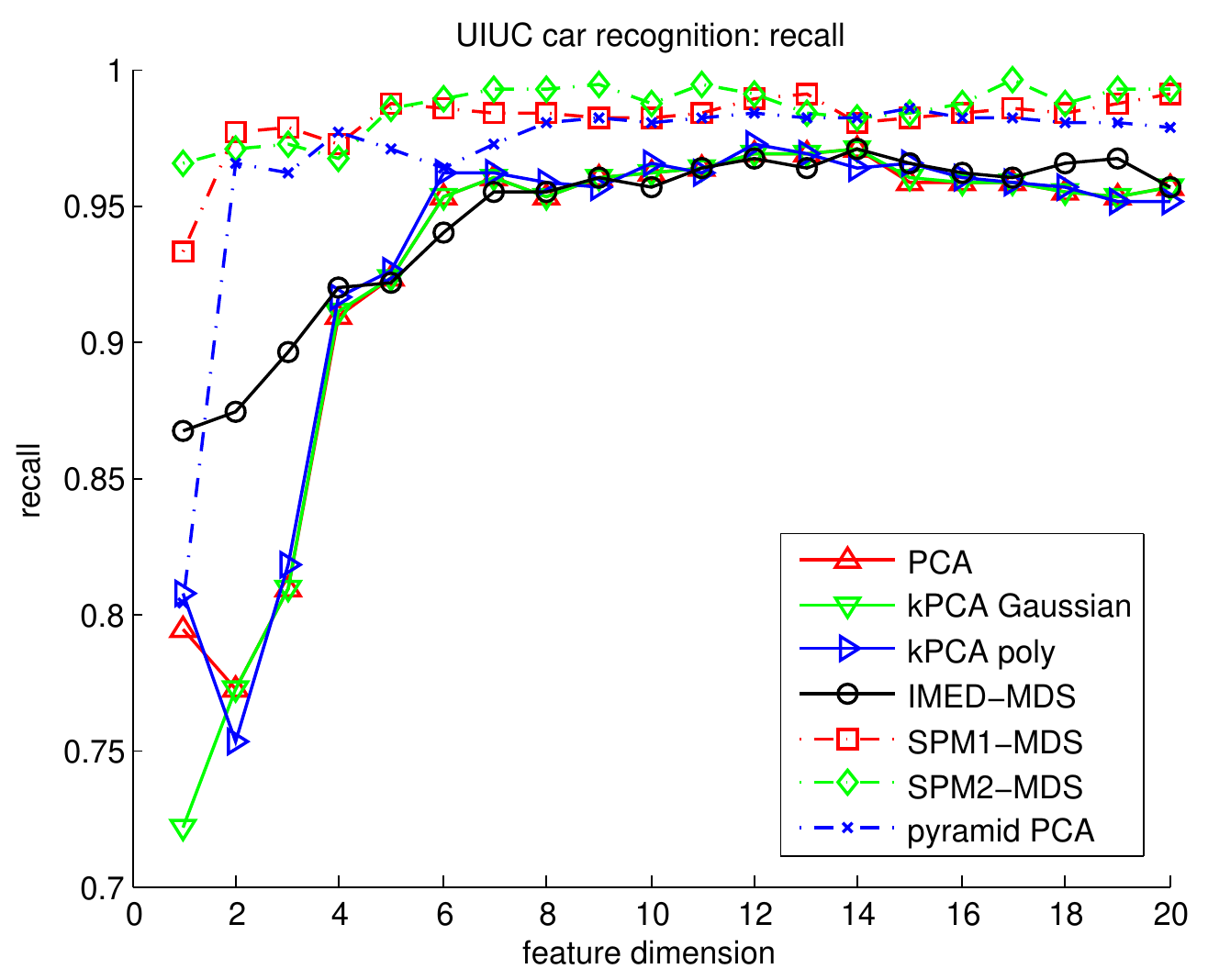}}
    \subfloat[accuracy]
      {\label{fig:UIUC_accuracy}
       \includegraphics[width=0.32\textwidth]{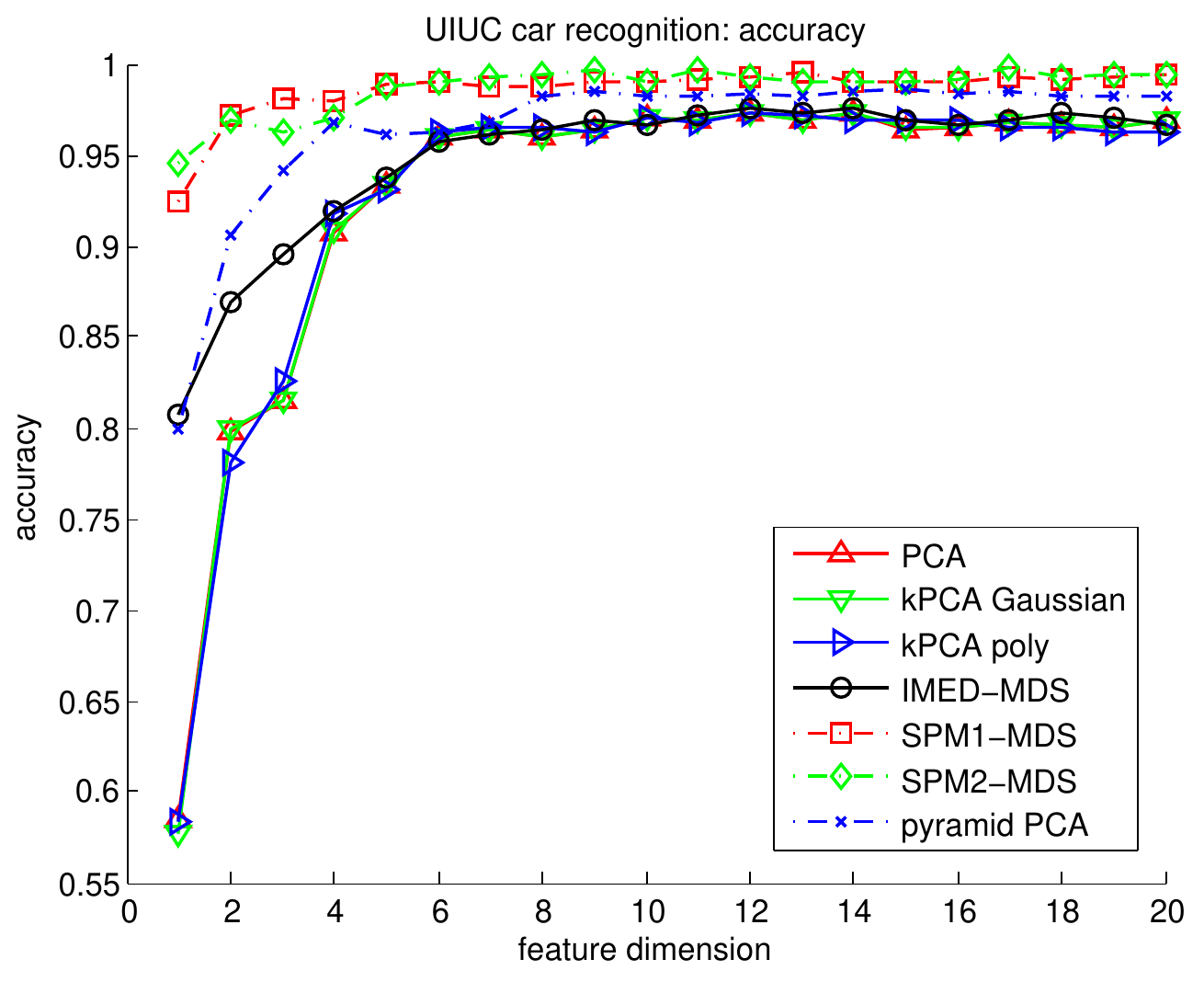}} 
  \caption{UIUC car recognition performance. }
  \label{fig:UIUC_car_performance}
\end{figure*}

\begin{figure*}
  \centering
    \subfloat[PCA]
      {\label{fig:scatter_PCA}
       \includegraphics[height=0.21\textwidth]{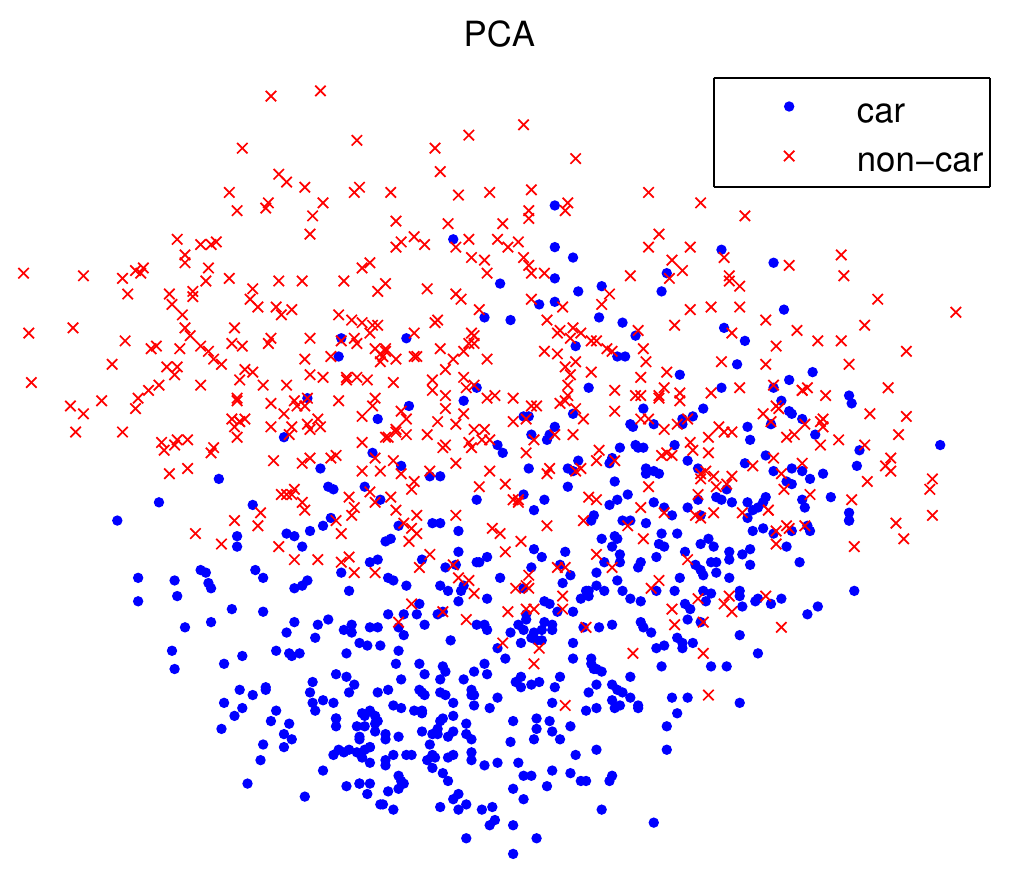}} 
    \subfloat[kPCA Gaussian]
      {\label{fig:scatter_kPCA_Gaussian}
       \includegraphics[height=0.21\textwidth]{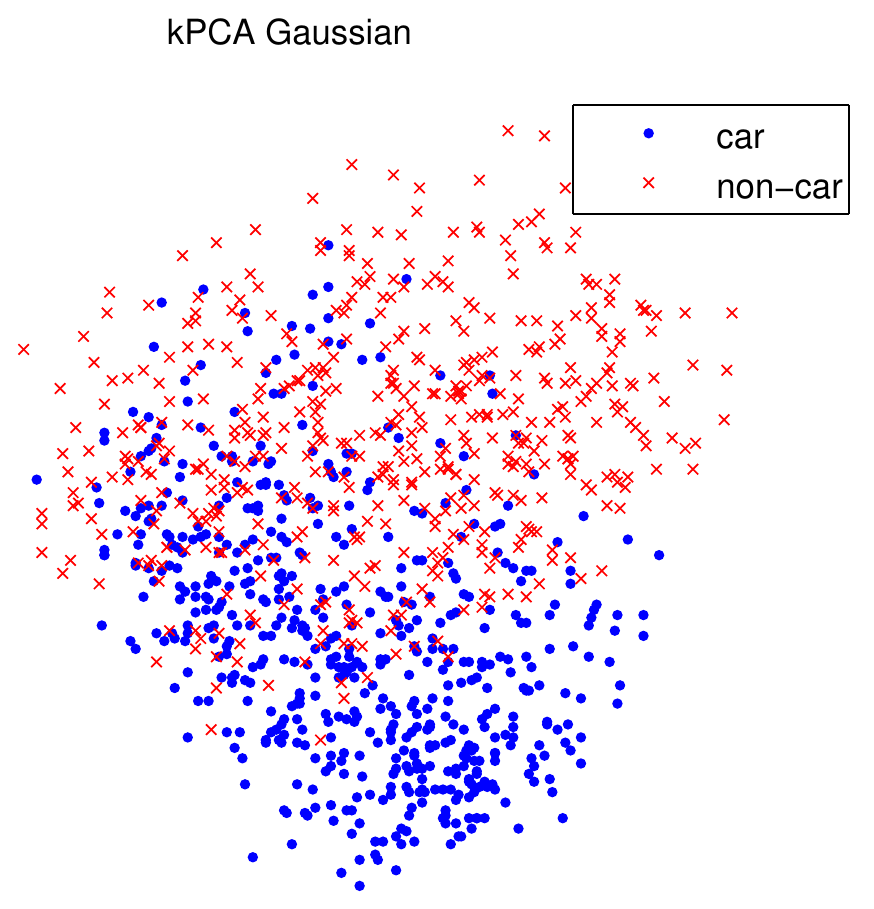}}
    \subfloat[kPCA poly]
      {\label{fig:scatter_kPCA_poly}
       \includegraphics[height=0.21\textwidth]{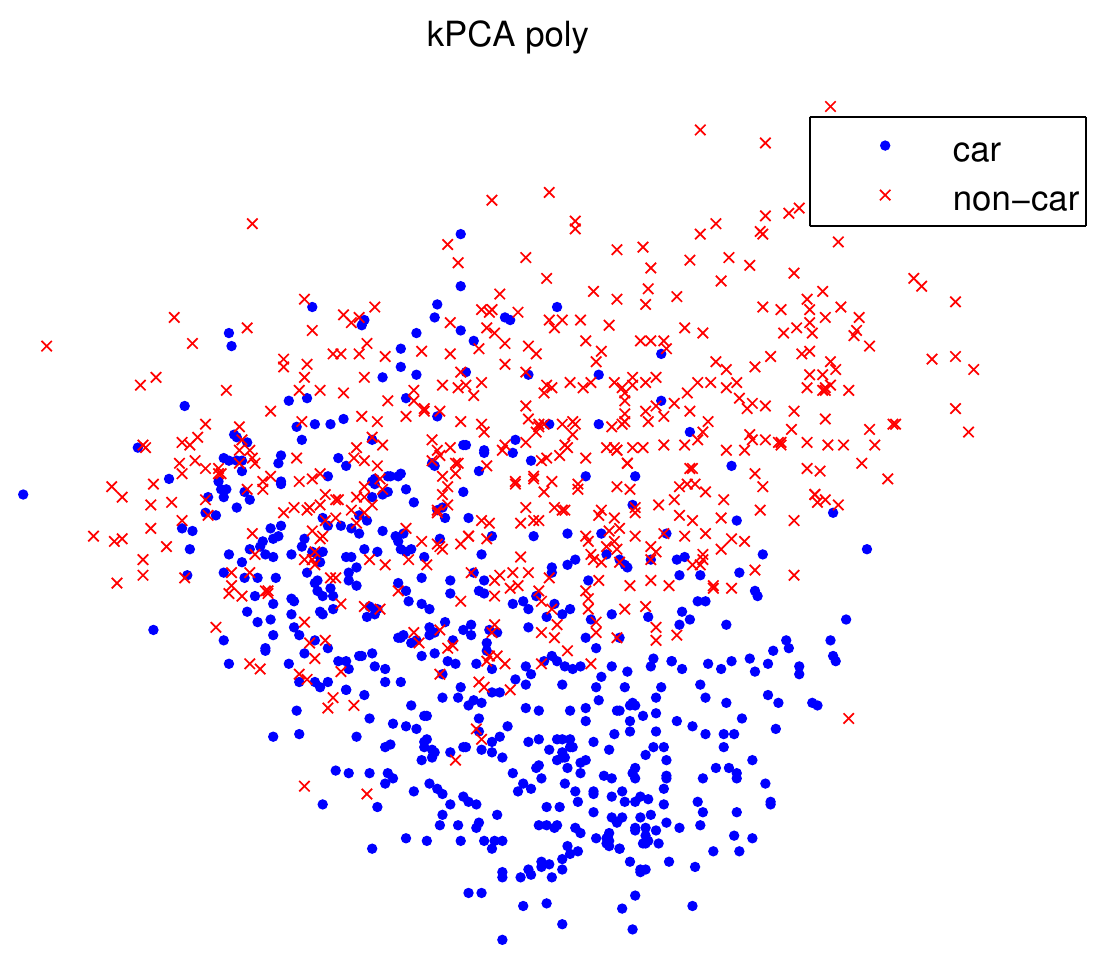}} 
    \subfloat[IMED-MDS]
      {\label{fig:scatter_IMED-MDS}
       \includegraphics[height=0.21\textwidth]{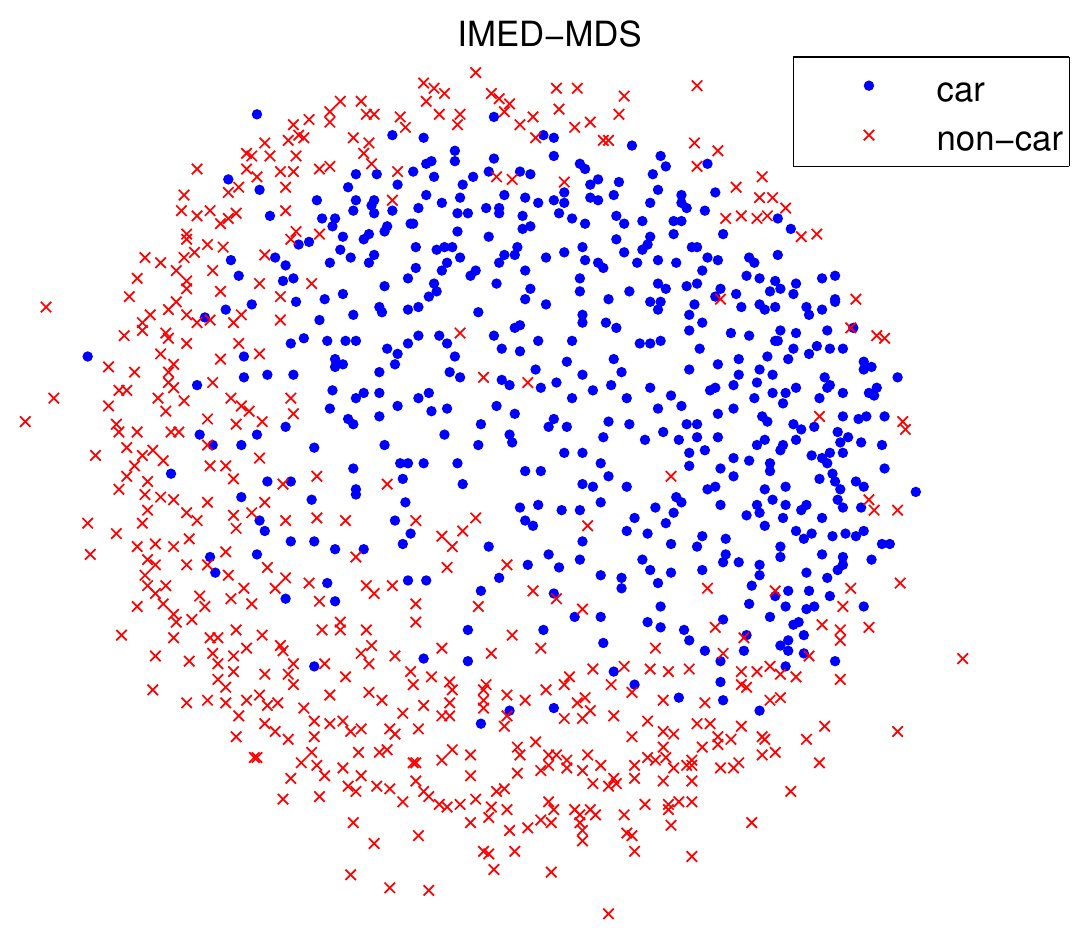}}
    \\
    \subfloat[SPM1-MDS]
      {\label{fig:scatter_SPM1-MDS}
       \includegraphics[height=0.21\textwidth]{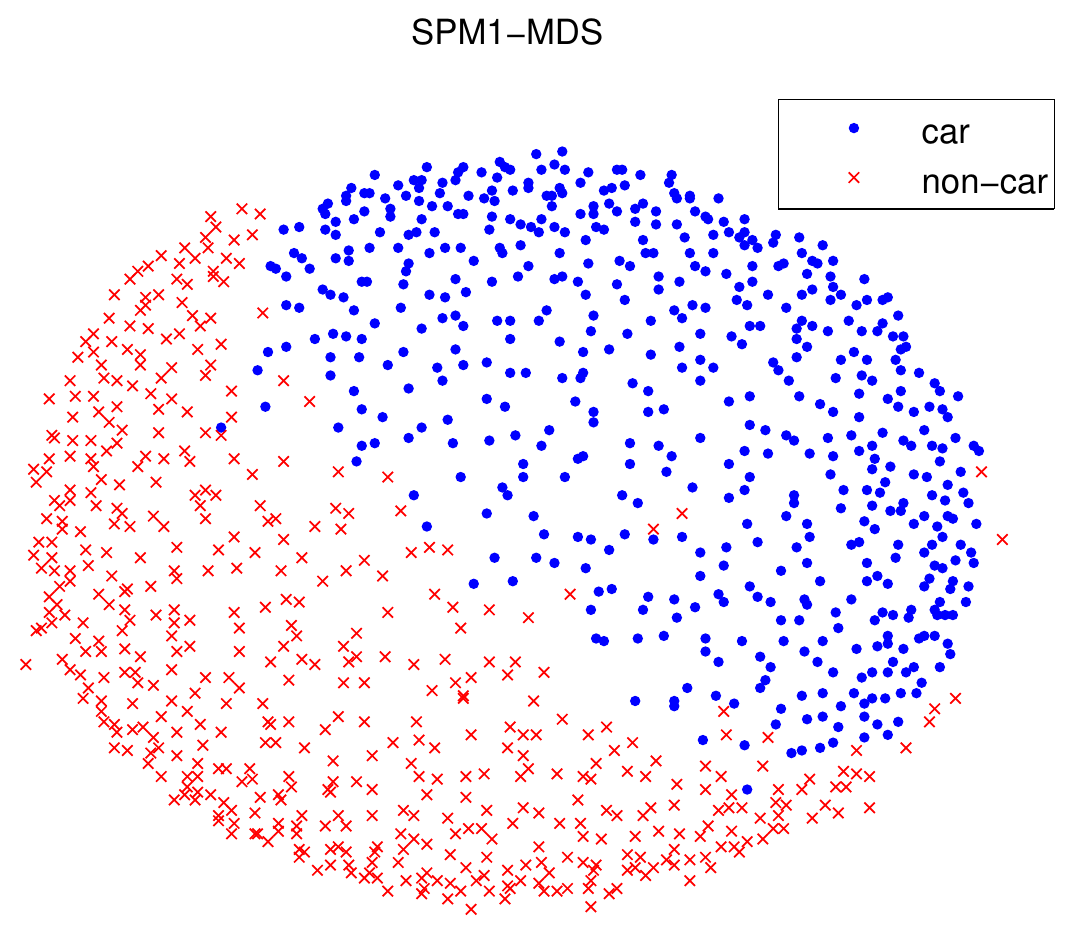}}
    \subfloat[SPM2-MDS]
      {\label{fig:scatter_SPM2-MDS}
       \includegraphics[height=0.21\textwidth]{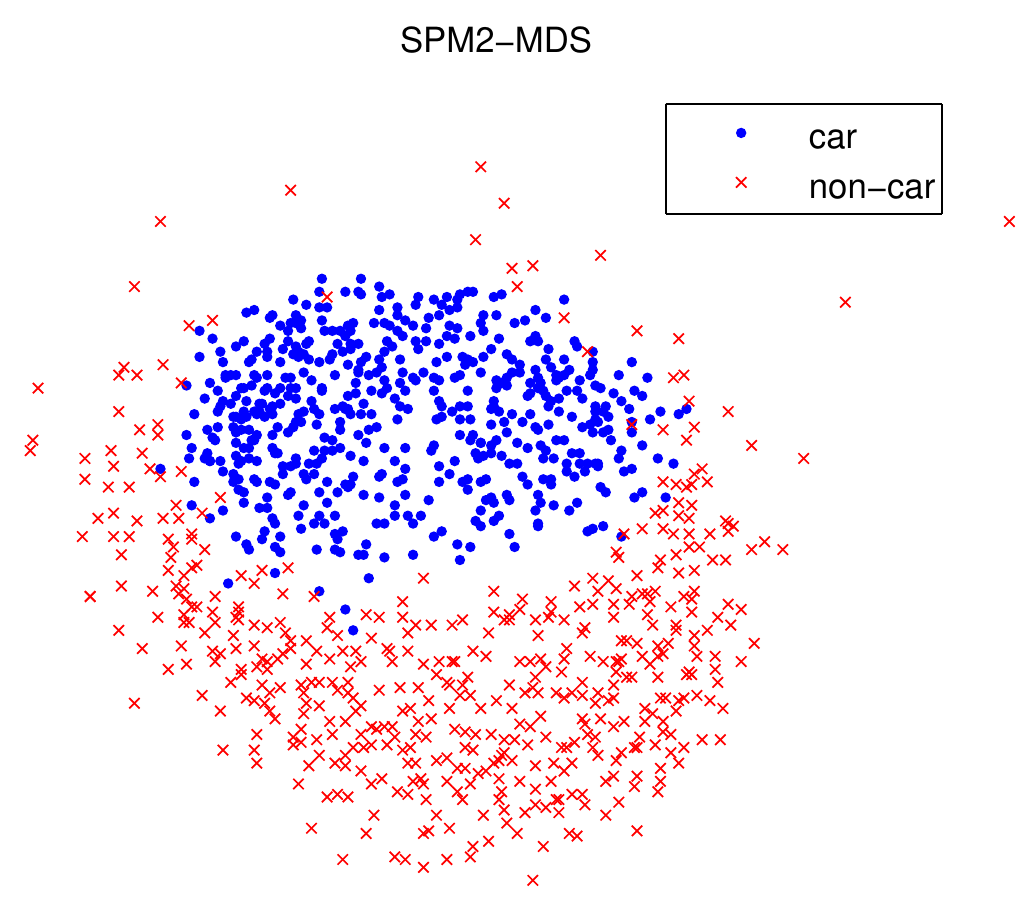}} 
    \subfloat[pyramid PCA]
      {\label{fig:scatter_pyramidPCA}
       \includegraphics[height=0.21\textwidth]{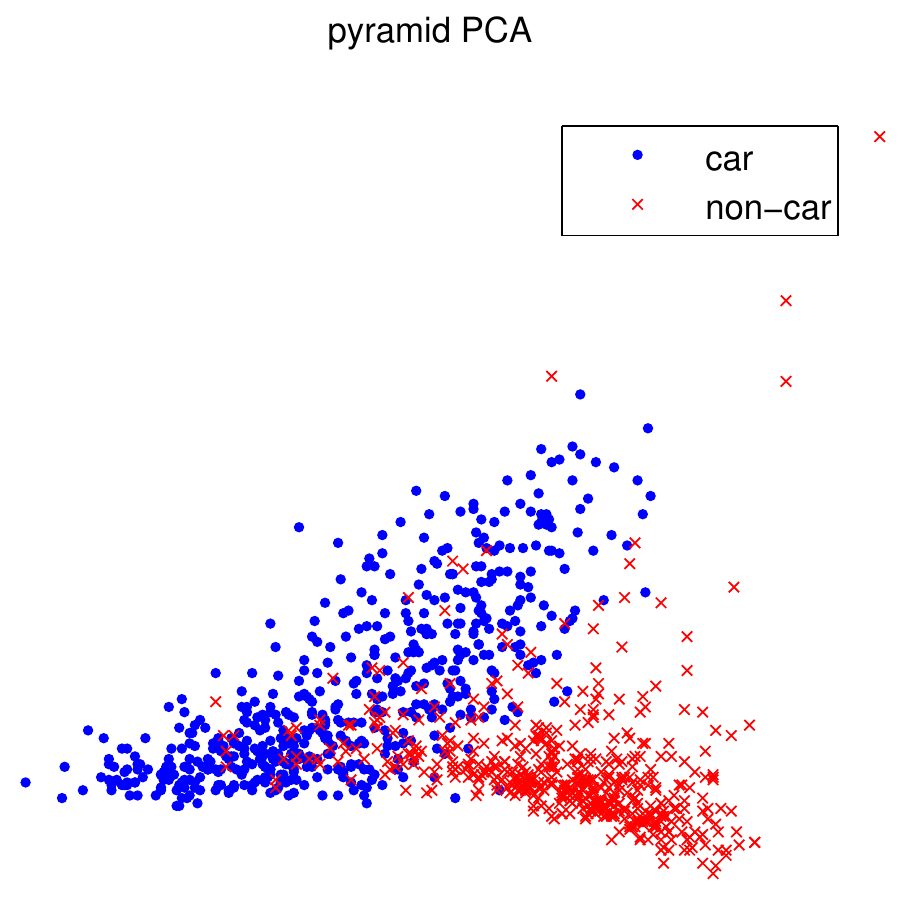}}
  \caption{2-d feature scatter plots of UIUC car image dataset with different features.  }
  \label{fig:car_scatter}
\end{figure*}

After we have obtained the fixed-length features of all images, we use the features of training images to learn a binary RBF kernel SVM \cite{SVM, libsvm}, and use it to classify the features of testing images. Each dimension of the feature vector is normalized to 0-mean and unit standard deviation. In the radial basis function $\exp(-||\mathbf{u}-\mathbf{v}||^2/\gamma)$, we set $\gamma$ as the feature vector length. 
The experiment is repeated for different feature vector lengths from 1 to 20. We show the precision, recall and accuracy in Fig. \ref{fig:UIUC_car_performance}. 
We also provide the feature scatter plots of different methods for feature length $m=2$ in Fig. \ref{fig:car_scatter}. 

In Fig. \ref{fig:UIUC_car_performance}, we can observe that IMED-MDS method performs slightly but not significantly better than directly applying PCA or kernel PCA on raw gray-level intensities, and the superiority of IMED-MDS is more obvious when feature dimension is low. 
Spatial pyramid based methods do perform much better than other methods. Especially, SPM1-MDS and SPM2-MDS methods outperform all other methods, including pyramid PCA, at all feature dimensions. While the precision and recall of PCA, kernel PCA and IMED-MDS methods saturate at $98\%$ and $96\%$ respectively, the precision and recall of SPM1-MDS and SPM2-MDS saturate at $100\%$ and $99\%$ respectively. At low feature dimensions ($m\leq5$), the accuracy of PCA and kernel PCA are very low, but the SPM1-MDS and SPM2-MDS perform almost as equally well as at very high dimensions. 

In Fig. \ref{fig:car_scatter}, we can also see that SPM1-MDS and SPM2-MDS separate car and non-car images with very clear class boundary curves in 2-d feature space. 


\section{Conclusions and Future Work}
In this paper, we have presented a feature learning framework by combining multidimensional scaling with image distance measurement, and compared it with a number of popular existing feature extraction techniques. To the best of our knowledge, we are the first to explore MDS on image distances such as IMage Euclidean Distance (IMED) and Spatial Pyramid Matching (SPM) distance. 

We have introduced a unified framework for both MDS model training and new data encoding based on the standard Levenberg-Marquardt algorithm. Our two-stage iterated Levenberg-Marquardt algorithm for MDS model training is an efficient solution, and has shown good running time performance compared with other off-the-shelf implementations (Fig. \ref{fig:ILMA_runningtime}). 

In the car recognition experiment, we have demonstrated the power of MDS features. 
MDS features learned from SPM distances achieve the best classification performance on all feature dimensions. 
The good performance of MDS features attributes to the semantics-sensitive image distance, since it captures very different information from the images than traditional feature extraction techniques. The MDS further embeds such information into a low-dimensional feature space, which also captures the inner structure of the entire dataset. The MDS embedding is a very necessary step, since in Fig. \ref{fig:UIUC_car_performance} 
we can see the performance of MDS features learned from SPM distances is significantly better than simply running PCA on spatial pyramid vectors. 

Our ongoing work on this method explores these directions: 
\begin{enumerate}
\item We study more image distance measurements, such as the Integrated Region Matching (IRM) distance, which was originally designed for semantics-sensitive image retrieval systems  \cite{simplicity}. Performance of MDS codes learned from such distances can be evaluated and compared with the SPM-MDS method in this paper. 

\item Our MDS feature learning framework can be validated on larger datasets with many categories of color images of different sizes. For example, we can validate the methodology on the popular Caltech-101  dataset \cite{caltech101} or the COREL dataset \cite{simplicity, corel-divided}. 

\item In Eq. (\ref{eq:encoding}), rather than using the entire training set, we can also use only a subset of the training images to encode new data. It would be interesting to see how the performance varies by applying different subset selection strategies and different sizes of the subset. 

\item Currently the two-stage iterated Levenberg-Marquardt algorithm is implemented in MATLAB\footnote{Code is available at \url{https://sites.google.com/site/mdsfeature/}}. We are  also recoding it in C/C++ with the lmfit library\footnote{\url{http://joachimwuttke.de/lmfit/}}, which will be more computationally efficient. 
\end{enumerate}


\bibliographystyle{IEEEtran}
\bibliography{example}

\begin{IEEEbiography}[{\includegraphics[width=0.8in,keepaspectratio]{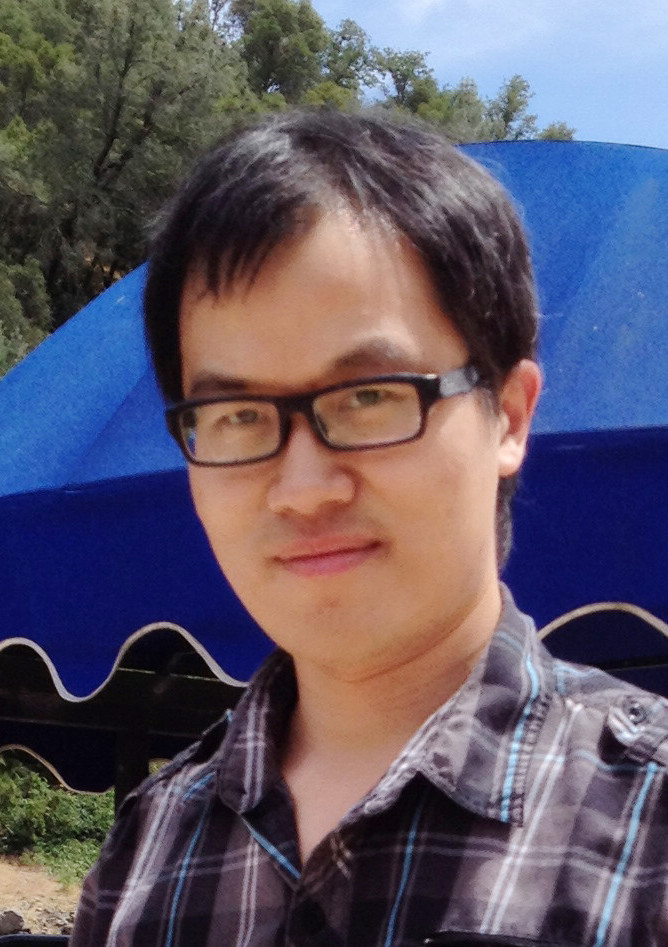}}]{Quan Wang}
Quan Wang is currently working towards his Ph.D. degree in Computer and Systems Engineering in the Department of Electrical, Computer, and Systems Engineering at Rensselaer Polytechnic Institute. He received a B.Eng. degree in Automation from Tsinghua University, Beijing, China in 2010. He worked as research intern at Siemens Corporate Research, Princeton, NJ and IBM Almaden Research Center, San Jose, CA in 2012 and 2013, respectively. 
His research interests 
include feature learning, medical image analysis, object tracking, content-based image retrieval and photographic composition.
\end{IEEEbiography}

\begin{IEEEbiography}[{\includegraphics[width=0.8in,keepaspectratio]{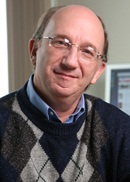}}]{Kim L. Boyer}
Dr. Kim L. Boyer is currently Head of the Department of Electrical, Computer, and Systems Engineering at Rensselaer Polytechnic Institute. He received the BSEE (with distinction), MSEE, and Ph.D. degrees, all in electrical engineering, from Purdue University in 1976, 1977, and 1986, respectively. 
From 1977 through 1981 he was with Bell Laboratories, Holmdel, NJ; from 1981 through 1983 he was with Comsat Laboratories, Clarksburg, MD. From 1986--2007 he was with the Department of Electrical and Computer Engineering, The Ohio State University. 
He is a Fellow of the IEEE, a Fellow of IAPR, a former IEEE Computer Society Distinguished Speaker, and currently the IAPR President. Dr. Boyer is also a National Academies Jefferson Science Fellow at the US Department of State, spending 2006--2007 as Senior Science Advisor to the Bureau of Western Hemisphere Affairs. He retains his Fellowship as a consultant on science and technology policy for the State Department. 
\end{IEEEbiography}

\end{document}